\documentclass[10pt,journal]{IEEEtran}
\usepackage{amsmath,amsfonts}
\usepackage{hyperref}
\usepackage{algorithmic}
\usepackage{algorithm}
\usepackage{array}
\usepackage[caption=false]{subfig}
\usepackage{textcomp}
\usepackage{stfloats}
\usepackage{url}
\usepackage{verbatim}
\usepackage{graphicx}
\usepackage{cite}
\usepackage{booktabs}
\usepackage{mathptmx}
\usepackage{multirow}

\usepackage{xcolor}
\newcommand{\blue}[1]{\textcolor{black}{#1}} %benutzt Mark
\newcommand{\bluee}[1]{\textcolor{black}{#1}} %benutzt Mark

\begin{document}

\title{On-Device Training of Fully Quantized Deep Neural Networks on Cortex-M Microcontrollers}

%\author{Anonymous
        % <-this % stops a space
%\thanks{This paper was produced by the IEEE Publication Technology Group. They are in Piscataway, NJ.}% <-this % stops a space
%\thanks{Manuscript received -; revised -.}}
\author{Mark Deutel$^{1,2}$, Frank Hannig$^{1}$, Christopher Mutschler$^{2}$, and Jürgen Teich$^{1}$\thanks{$^{1}$ Friedrich-Alexander-Universität Erlangen-Nürnberg (FAU)}\thanks{$^{2}$ Fraunhofer IIS, Fraunhofer Institute for Integrated Circuits IIS}}

% The paper headers
%\markboth{International Conference on Hardware/Software Codesign and System Synthesis}%
%{Shell \MakeLowercase{\textit{et al.}}: A Sample Article Using IEEEtran.cls for IEEE Journals}

%\IEEEpubid{0000--0000/00\$00.00~\copyright~2024 IEEE}
% Remember, if you use this you must call \IEEEpubidadjcol in the second
% column for its text to clear the IEEEpubid mark.

\maketitle

\begin{abstract}
On-device training of DNNs allows models to adapt and fine-tune to newly collected data or changing domains while deployed on microcontroller units (MCUs). However, DNN training is a resource-intensive task, making the implementation and execution of DNN training algorithms on MCUs challenging due to low processor speeds, constrained throughput, limited floating-point support, and memory constraints. In this work, we explore on-device training DNNs for different sized Cortex-M MCUs (Cortex-M0+, Cortex-M4, and Cortex-M7). We present a method that enables efficient training of DNNs completely in place on the MCU using fully quantized training (FQT) and dynamic partial gradient updates. We demonstrate the feasibility of our approach on multiple vision and time-series datasets and provide insights into the tradeoff between training accuracy, memory overhead, energy, and latency on real hardware. The results show that compared to related work, our approach requires 34.8\% less memory and has a 49.0\% lower latency per training sample, with dynamic partial gradient updates allowing a speedup of up to 8.7 compared to fully updating all weights.
\end{abstract}


\section{Introduction}
\label{sec:introduction}
\IEEEPARstart{A}{dapting}, fine-tuning, or retraining deep neural networks (DNNs) on microcontroller units (MCUs) to improve alongside a changing input domain or application is not a straightforward task~\cite{lin2023tiny}. Typically, once the DNN is deployed on the MCU, its trainable weights are stored read-only in Flash, and adapting the DNN without reprogramming the MCU is often not feasible. In addition, the limited memory and computational resources on the MCU require that the weights and feature maps of a DNN are often stored in quantized form, which further complicates regular DNN training normally performed in floating-point. Moreover, DNN training itself adds significant memory and computational overhead, making on-device training on MCUs for DNNs of meaningfull size and complexity even more challenging.

However, training DNNs on MCUs has a number of highly desirable properties, such as (a) a reduced communication of the MCU with remote training servers, since no data recorded by the MCU needs to be sent away for offline training, (b) increased privacy, since training with data from the input domain, which may contain personally identifiable information such as speech or likeness, can be performed directly on the MCU without any persistent storage of data on a remote server, (c) energy efficiency, since MCUs have an energy footprint several orders of magnitude smaller than any GPU-based server, and (d) in place training with zero downtime, since DNN training can be performed on the fly alongside the regular inference of recorded samples on the MCU.

\subsection{Challenges}
\label{sec:intro:challanges}

Supervised training of DNNs relies on two algorithms: (a) backpropagation (BP), which is used to compute local gradients for each intermediate layer of the DNN based on the chain rule, and (b) stochastic gradient descent (SGD), which is used to update the trainable weights of each layer based on the local gradients calculated by BP. Usually, the gradients used by SGD are stochastically approximated by accumulating them over a small subset of samples (minibatches) of all available training samples. The combination of BP and SGD is well-known, versatile, and the backbone of all major AI advances of recent years. However, it is also extremely memory- and compute-intense, resulting in DNN training usually being performed on larger server systems and with computing accelerator cards like GPUs. This results in major challenges that need to be addressed to enable DNN training on regular MCUs.

The high computational cost of DNN training lies in the "backward pass", i.e., the BP and SGD algorithms, that must be executed for each training sample in addition to the "forward pass", i.e., the inference. The backward pass is responsible for propagating the error computed for the last layer of the DNN by a loss function to calculate local errors for all intermediate layers. As a result, for each layer, at least one partial derivative with respect to its input during the forward pass must be computed, while for all trainable layers another partial derivative with respect to their weights must also be computed to facilitate the parameter updates performed by SGD. For example, for linear layers and when using minibatches, this results in two additional transposed matrix multiplications, while for convolutional layers, two additional "transposed convolutions" have to be performed in addition to the forward pass. In conclusion, for trainable layers in a DNN, the added computational cost of performing their backward pass is about twice as high as the computational cost of performing their forward pass.

\begin{figure}
    \centering
    \includegraphics[width=.7\linewidth, clip, trim=0.0cm 1.8cm 1.5cm 0.0cm]{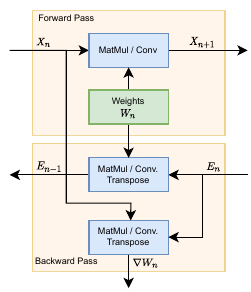}
    \caption{Data dependencies between forward and backward pass of linear and convolutional layers in a DNN.}
    \label{fig:overview_bp}
\end{figure}

The high memory overhead of DNN training is the result of four aspects: First, the partial derivatives of most layers commonly used in DNNs depend on intermediate tensors used in the forward pass of the network, see Fig.~\ref{fig:overview_bp} for a schematic overview. If only inference is performed, these tensors are usually short-lived because they only temporarily store data between successive layers.
Therefore, to save RAM, many inference libraries for MCUs try to optimize memory requirements by reusing areas on the heap for multiple tensors during an inference pass ~\cite{lin2020mcunet, deutel2023energy}.
\IEEEpubidadjcol
However, when performing backpropagation, many of these intermediate results need to be kept in memory longer, since they are required by operations in the backward pass that can only be executed after the complete forward pass has finished and the backward pass has propagated the error back to the corresponding layer. As a result, optimization of heap usage becomes less efficient.

Second, when performing SGD, many inference frameworks process an entire minibatch of samples at once, adding another "batch dimension" to the input, output, and all intermediate tensors, and thus greatly increasing the memory utilization, which now additionally depends on the minibatch size used.

\bluee{Third, additional memory must be provided to store training data over time, either as a labeled dataset for supervised training or a replay buffer for continual learning. On an MCU, this typically requires external memory from which data can be swapped in and out of RAM.}

Fourth, since the weights of all trained layers are dynamically updated at runtime, i.e., they are not read-only, they cannot be stored in Flash, where most MCU inference libraries would place them, but must instead be placed in RAM.

\subsection{Contribution}
\label{sec:contribution}

To address the challenges outlined in Section ~\ref{sec:intro:challanges} and to make on-device training on tiny devices such as Cortex-M MCUs feasible, we propose the following contributions in this paper: 

We introduce (i) an 8-bit fully quantized training (FQT) algorithm that quantizes both weights and intermediate feature tensors during the forward pass and errors and gradients tensors during the backward pass by using the same linear quantization scheme also used for regular DNN inference. As a result, our approach enables on-device DNN training on MCUs online, in place, and without any transformations, changes, or conversions to the DNN's representation in memory or code between inference and training. 

Furthermore, to allow FQT while still being able to achieve stable training results similar to regular floating-point training on GPUs, we propose (ii) a memory-efficient version of minibatching with standardized gradients as well as (iii) a method to optimize the distribution of values in the 8-bit range of quantized tensors by adapting the quantization parameters of trained weight tensors dynamically during training, similar to quantization-aware training (QAT)~\cite{jacob2018quantization}. 

To further reduce the computational overhead of our approach, we employ (iv) a dynamic gradient update strategy that is fully compatible with our FQT method, which prunes the computational tree of BP per training sample based on an online computed magnitude of error based heuristic. 

Finally, we demonstrate the flexibility of our approach for both transfer learning and full DNN training for multiple datasets (11 in total), and provide insights into the trade-off between memory, latency, energy, and DNN accuracy for three Cortex-M-based MCUs, see Tab.~\ref{tab:mcus}.

The rest of this paper is structured as follows. Section~\ref{sec:background} outlines work on on-device training of DNNs on MCUs related to ours. In Section~\ref{sec:implementation} we discuss our approach and its implementation in more detail. In Section~\ref{sec:evaluation} we present our evaluation and then conclude in Section~\ref{sec:conclusion}.
\section{Related Works}
\label{sec:background}

On-device training on MCUs is gaining popularity as it promises a much simpler and more energy-efficient solution for DNN fine-tuning and adaptation than training on a remote server and then redeploying the model. As a result, much research on on-device training so far has focused on achieving a better trade-off between memory overhead, computing resource requirements, and accuracy. 

A popular approach to reduce memory utilization is to recompute intermediate results from the forward pass, which are required by the backward pass, instead of keeping them in memory~\cite{gruslys2016memory, chen2016training}. While this helps to minimize the additional memory overhead introduced by training, it does so at the cost of a significant additional computational overhead due to multiple computations of results. Another proposed approach is to reduce the memory footprint of intermediate activation tensors by constructing a dynamic and sparse computational graph~\cite{liu2019dynamic}. However, the approach as presented by the authors is only designed to minimize activations of the forward pass and still uses a matrix-vector product, albeit reduced in size, to decide which part of the computational graph to eliminate. Other approaches focus on a hierarchical approach with different levels of granularity at which to drop computations, such as the data, model, and algorithm levels~\cite{wang2019e2}. Finally, reducing memory utilization during both inference and training by processing both activations and weights quantized has also been proposed~\cite{wang2018training, lin2022device}.

Due to the limited memory and computational resources on embedded devices, the most common DNN training tasks addressed by researchers so far are transfer learning and fine-tuning tasks, where an already well-trained DNN is adapted to a changed input domain or application. The reason for the popularity of these tasks is that they usually do not require training of a DNN from scratch, but only fine-tuning of its last layers. For example, Tiny-transfer-learning~\cite{cai2020tinytl} freezes the weights of the DNN and trains only the biases, for which updates can be computed much easier and with less memory overhead than for weights. A similar approach is to train only batch normalization layers~\cite{frankle2020training, mudrakarta2018k}. Alternatively, TinyOL~\cite{ren2021tinyol} trains only the last layer of the DNN to achieve reduced memory and computational overhead, while \cite{mudrakarta2018k} introduces small additional "patches" of trainable weights to a well-trained DNN. However, while training only subsets of a DNN's weights can lead to adequate training results, especially for more complex transfer learning tasks or larger domain shifts, the results can be poor~\cite{cai2020tinytl}, making these strategies not as reliable as full DNN training.

An approach to enable full DNN training on MCUs is described in~\cite{patil2022poet}, which uses two techniques: first, freeing up memory for activation tensors early at the cost of having to recompute them later, and second, paging where parts of the activation tensors are swapped out to a secondary, larger memory. However, swapping to external memory, in particular, is slow and makes certain assumptions about the MCU that may now be applicable in all cases. 

Alternatively, the authors in~\cite{lin2022device} use a combination of quantized training, which they call quantization-aware scaling, combined with updating only parts of the weights on-device at runtime. This approach is probably the closest to ours, although there are some concerns that we try to address in this paper. 
First, the approach described by the authors seems to be optimized exclusively for their own DNN architecture MCUNet~\cite{lin2020mcunet}, while our method focuses on general CNN support for different domains and applications. 
Second, the authors mostly provide simulated results for transfer-learning using their on-device training approach, while our work aims to explore the trade-off between performance and resource requirements on MCUs of different scale (Cortex-M0+, Cortex-M4, and Cortex-M7) and for different tasks. 
Third, the gradient selection method proposed by the authors for their on-device training is performed entirely at compile time (offline). This effectively freezes parts of the DNN at runtime, which, similar to the concerns we mentioned about transfer learning, can lead to poor training results for certain datasets and domains. \blue{To overcome this problem, our work proposes a method to dynamically evaluate which gradients to compute for each sample online during on-device training.}

Finally, several memory-aware on-device learning frameworks have been proposed for larger embedded systems, i.e., mobile and IoT edge platforms~\cite{wang2022melon, xu2022mandheling, gim2022memory, qi2018enabling}. While inspiring, the methodologies proposed and implemented by these frameworks are not directly applicable to Cortex-M MCUs, mainly due to the much more constrained memory and compute resources present on MCUs compared to larger mobile and edge platforms.
\section{Concepts}
\label{sec:implementation}

To make supervised training feasible on MCUs, we propose a training framework centered around two techniques that we will discuss below: Fully quantized training (FQT), see section ~\ref{sec:implementation:quant_bp}, and dynamic sparse gradient updates, see section ~\ref{sec:implementation:dyn_grad}.

\subsection{Fully Quantized Training}
\label{sec:implementation:quant_bp}

\begin{figure*}
    \centering
    \subfloat[Quantization Aware Training (QAT)]{\label{fig:impl:qat_vs_fqt:qat}%
        \includegraphics[width=0.52\textwidth]{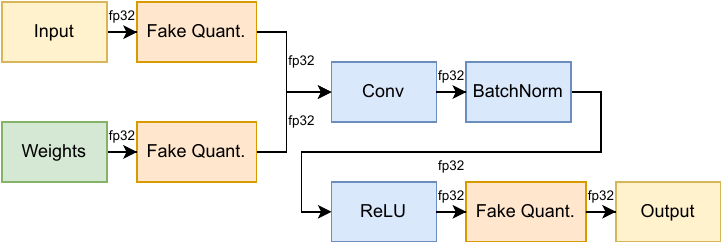}}\hfill
    \subfloat[Fully Quantized Training (FQT)]{\label{fig:impl:qat_vs_fqt:fqt}%
        \includegraphics[width=0.42\textwidth]{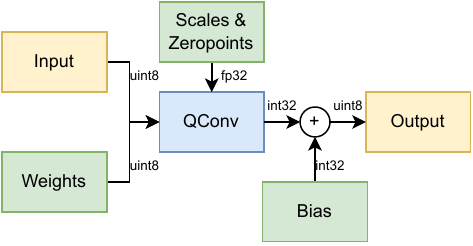}}
    \caption{Schematic comparison of a convolutional block with ReLU activation and Batchnorm for Quantization Aware Training (QAT) and Fully Quantized Training (FQT). In FQT the convoltional, Batchnorm and ReLU layers have been folded into a single monolithic QConv layer.}
    \label{fig:impl:qat_vs_fqt}
\end{figure*}

Typically, DNNs deployed on a MCU for inference are quantized using a linear quantization. This is done to save memory and because integer arithmetic can be performed more efficiently on most MCUs than floating-point arithmetic. Usually, quantization is implemented per tensor in the form of $v_q = \lfloor\frac{v_{f}}{s}\rfloor + z$, where $s$ and $z$ are two parameters \textit{scale} and \textit{zero point} derived from the distribution of floating-point values in the original unquantized tensor. Therefore, in order to train a DNN in place without any conversion, training must be performed using the same quantization scheme already established for inference. In addition, this has the advantage that all optimizations already established for quantized inference can be reused for training.
%First, because training can be done in place without having to change or transform the deployed model, and second, because all the optimizations already established by quantization for inference can be reused. 

A well-established method for training quantized DNNs on GPU-based systems is called quantization-aware training (QAT)~\cite{jacob2018quantization}. However, while the method helps to train DNNs robust to additional errors introduced by quantization, the DNN is not actually trained quantized, but still processed in floating-point arithmetic, with quantization only simulated, see Fig.~\ref{fig:impl:qat_vs_fqt:qat} for a schematic overview. As a result, in QAT, the DNN is not fully transferred to quantized space until training is complete. Therefore, QAT does not provide any immediate memory or computational savings for the training process itself. In addition, using QAT would require a costly conversion process at the beginning and end of training.

To overcome the problems of QAT, and to achieve our goal of in place, online training on fully quantized DNNs, we propose a training method based on FQT, facilitating training directly with the quantized weights of a DNN and using the already established quantization scheme for inference, see Fig.~\ref{fig:impl:qat_vs_fqt:fqt}. As a result, our method is flexible, since there is no difference in how the DNN is represented in memory and code between inference and training. Therefore, the same forward pass of a DNN executed on-device using our method can be used for both regular inference and training at the same time. Furthermore, since both the forward and backward pass of our training method are performed on quantized weights, feature maps, errors and gradients, the method is significantly more memory-efficient than traditional floating-point based training.

BP for DNNs is based on the chain rule, and calculation of local errors and gradients for both linear and convolutional layers can be described as 
\begin{align}
E_{n-1}&=W_n^{T} \cdot E_n \label{eq:backward_pass_err} \\
\nabla W_n&=E_{n} \cdot X_n^{T} \label{eq:backward_pass_weight}
\end{align}
with the $\cdot$ operator either being a matrix multiplication or a 2D-convolution when using minibatches. Therefore, Eq.~\eqref{eq:backward_pass_err} calculates the error $E_{n-1}$ of the $n-1$th layer of the DNN, given the error $E_{n}$ and weight $W_n$ of the $n$th layer, and Eq.~\eqref{eq:backward_pass_weight} calculates the gradient $\nabla W_n$ used to update the weights of the $n$th layer given the error $E_{n}$ and the input of the $n$th layer $X_n$, see also Fig.~\ref{fig:overview_bp}.
Note that both Eq.~\eqref{eq:backward_pass_err} and~\eqref{eq:backward_pass_weight} are based on the same mathematical operation $\cdot$ performed during the forward pass of the  layer
\begin{equation}
Y_n=W_n \cdot X_n \label{eq:forward_pass}
\end{equation}
with the addition of a transpose operator. In fact, Eq.~\eqref{eq:backward_pass_err} and~\eqref{eq:backward_pass_weight} are the partial derivatives of Eq.~\eqref{eq:forward_pass} with respect to $X_n$ and $W_n$ respectively. As a result, quantizing backpropagation is straightforward, since the same quantization scheme and optimizations that are used in the forward pass during inference can be directly applied again. Therefore, for example, the fully quantized version of Eq.~\eqref{eq:backward_pass_err} for each element $e_{n-1} \in E_{n-1}$ can be computed from $e_{n} \in E_{n}$ and $w_n \in W_n$ as 
\begin{equation}
\label{eq:backward_pass_err_quant}
e_{n-1}= \biggl\lfloor\frac{s_{w_n}s_{e_n}}{s_{e_{n-1}}}\sum (w_n - z_{w_n})(e_n - z_{e_n})\biggr\rfloor + z_{e_{n-1}}
\end{equation}
where $s_{w_n}$, $s_{e_n}$ and $s_{e_{n-1}}$ as well as $z_{w_n}$, $z_{e_n}$, and $z_{e_{n-1}}$ are the respective scale and zero point parameters for the quantized tensors $E_n$, $E_{n-1}$, and $W_n$. The quantized version of Eq.~\eqref{eq:backward_pass_weight} looks identical, however in our appraoch we omit the re-quantization of $\nabla W_n$ since we perform gradient descent locally in floating-point space in the next step.

For updating each element of the quantized weight $W_n$ of a given layer $n$ from a current state $i$ to a next state $i+1$ at the end of each mini-batch we use the gradient descent algorithm
\begin{equation}
w_{i+1} = \frac{1}{s_{w_{i+1}}} \left[ \left( w_i - z_{w_i} \right) s_{w_i}  - \ell \nabla w_i \right] + z_{w_{i+1}}
\label{eq:sgd_quant}
\end{equation}
adapted for quantization where $\ell$ is the learning rate, $s_{w_i}$ and $z_{w_i}$ are the current weight tensor's scale and zero point parameters, and $s_{w_{i+1}}$ and $z_{w_{i+1}}$ are the updated weight tensor's parameters. To update the scale and zero point parameters alongside the trainable weights, $s_{w_{i+1}}$, see Eq.~\eqref{eq:new_scale}, and $z_{w_{i+1}}$, see Eq.~\eqref{eq:new_zp}, are calculated based on the maximum and minimum values $f_{min}$ and $f_{max}$ observed in the intermediate results from $\left( w_i - z_{w_i} \right) s_{w_i}l\nabla w_i$ which are in floating-point space.
\begin{align}
s_{w_{i+1}} &= \frac{f_{max} - f_{min}}{255} \label{eq:new_scale} \\
z_{w_{i+1}} &= \biggl\lfloor -\frac{f_{min}}{s_{w_{i+1}}} \biggr\rfloor \label{eq:new_zp}
\end{align}

\bluee{To implement stochastic gradient descent (SGD), i.e., mini-batches, there are two viable approaches: (a) have all $b$ input samples of a mini-batch in RAM at the same time and process them all at once during a single network pass by introducing an additional dimension to the input, output, and all intermediate tensors, or (b) accumulate $\nabla W_n$ in a buffer over $b$ successive training steps of individual samples. Both options increase the amount of memory required compared to performing gradient descent on each sample, i.e., batch size $b=1$. However, not using SGD and instead relying on per-sample updates can have a significant impact on the training performance of the DNN, as negative training effects such as catastrophic forgetting~\cite{mccloskey1989catastrophic} become much more prevalent.}

\bluee{We have chosen version (b) for our SGD approach. The reason is that by introducing additional gradient buffers to accumulate $\nabla W_n$, the additional SRAM utilization to support SGD is independent of both the batch size $b$ and the full network depth $N$, and instead depends only on the number of layers actually trained in the DNN. Most importantly, this approach does not require all the samples in a minibatch to be stored in RAM at the same time. Furthermore, to minimize the additional memory overhead introduced by SGD, we do not consider extensions such as (Nesterov-) momentum or adaptive learning rate algorithms such as Adam~\cite{kingma2014adam}, as these techniques, while generally improving convergence and stability of training, come at the cost of significantly higher memory requirements due to the need to store multiple versions of the gradient buffer.}

Finally, while testing with quantized backpropagation and SGD, we found that standardizing the accumulated gradients per channel before performing the update step, similar to the intuition of RMSProp~\cite{hinton2012neural}, improves the convergence stability of our FQT approach. This results in a final per-element update step of our SGD approach in the form of

\begin{equation}
    w_{i+1} = \frac{1}{s_{w_{i+1}}} \left[ \left( w_i - z_{w_i} \right) s_{w_i}  - \ell \frac{\nabla w_i - \mu_{\nabla W}}{\sigma_{\nabla W}} \right] + z_{w_{i+1}}
\end{equation}

where $\mu_{\nabla W}$ and $\sigma_{\nabla W}$ are calculated as running mean and standard deviation of the local gradients $\nabla W$ calculated per sample.

\subsection{Dynamic Sparse Gradient Updates}
\label{sec:implementation:dyn_grad}

While quantized BP can help overcome the high memory overhead of BP, it does not directly contribute to reducing the algorithm's computational complexity\footnote{Note that from a practical standpoint, just using FQT often already reduces the latency per training sample, since many MCUs can compute more efficiently with integers than with floats}. Therefore, we propose \textit{Dynamic Sparse Gradient Updates} as a technique that can be used in conjunction with our FQT-based training method to reduce the computational overhead added by BP with minimal impact on the training performance of our method.

We base our approach to only partially update gradients during BP on the hypotheses that (a) during BP, the magnitude of values in the gradient tensors becomes smaller, (b) during BP, the distribution of high magnitude values becomes sparser for earlier layer, and (c) as training progresses and the DNN has gained a good understanding of the underlying optimization landscape, different samples will only activate smaller clusters of neurons in a DNN at once. As a result, as training progresses, more and more values in backpropagated error tensors will not be of high magnitude, i.e., they will not lead to a significant update of the corresponding weights. 

%We illustrate the effects discussed above in Fig.~\ref{fig:impl:heatmap}, where we visualize the distribution and absolute magnitude of the values of the gradient tensors of the last three linear layers for a given training sample of a DNN trained on the flowers dataset from our evaluation after the first and tenth training epochs. Based on this analysis, we argue that by monitoring the error tensors per training sample and per layer during BP and then eliminating values within the error tensors with an insignificant contribution, i.e. with an absolute magnitude close to zero, the computational complexity of BP can be significantly reduced without having a noticeable impact on the training performed.

\begin{figure}[t]
    \centering
    \includegraphics[width=.75\linewidth]{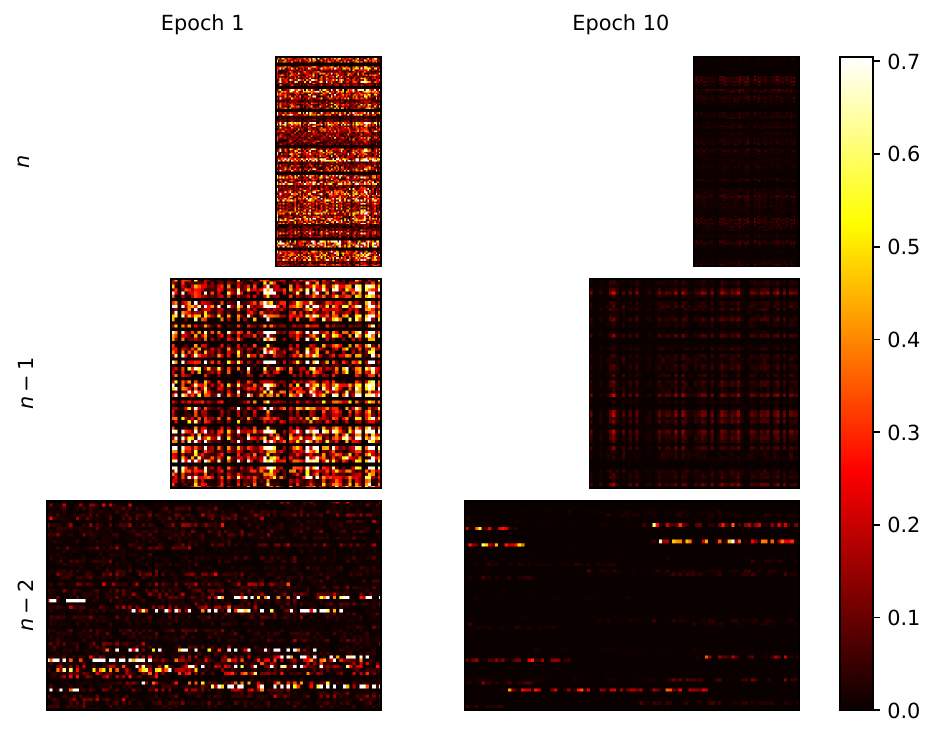}
    \caption{Heatmaps of the absolute values of the gradient tensors of the last three linear layers of a DNN trained on the flowers dataset, exemplarily for a training sample after the first epoch (left column) and the tenth epoch (right column) of training.}
    \label{fig:impl:heatmap}
\end{figure}

We illustrate the intuitions discussed above in Fig.~\ref{fig:impl:heatmap}, where we visualize the gradient tensors of the last three linear layers of a DNN trained on the flowers dataset from our evaluation. The heatmaps show the absolute values of the gradient updates of a single training sample after the first and the tenth epoch of training. We want to emphasize three aspects. First, the gradient tensors become sparser during the backward pass (compare rows $n$ and $n-2$). Therefore, our approach implements a per-layer gradient update rate. Second, in general, features are not learned randomly across neurons, but in structures that are activated at once, i.e., rows and columns for linear layers, features and channels for convolutional layers. \blue{This observation allows us to estimate the importance of parts of the gradient tensor, i.e., the magnitude of values, averaged per structure rather than per element, thereby decreasing the overhead of our method.}
Third, the overall magnitude of the gradient values decreases as training progresses (compare the two columns), resulting in fewer gradient structures worth updating in later stages of training than in early stages. As a result, we implement a dynamic update rate into our algorithm, which can decrease over the course of training.

Our proposed partial gradient updating algorithm monitors the absolute magnitude of the values in the error tensors of each layer for each training sample during BP as a simple heuristic to rank them. To limit the computational overhead of calculating the heuristic, our algorithm does not evaluate each value of the error tensors separately, but only observes the $l1$-norm per channel/column. This optimization is based on the assumption that trainable DNN layers represent features in structures and not randomly, so it is worthwhile to update the weights of an entire structure when a high error is observed. 

To calculate a per-layer update rate which can subsequently be used to select the total number of top-$k$ values that should be kept in an error tensor based on the magnitude of error ranking achieved before, our approach uses the following equation
\begin{equation}
k = \lfloor \min \{ \lambda_{min} + \lvert \varepsilon \rvert (\lambda_{max} - \lambda_{min}), 1 \} N \rfloor
\label{eq:part_bp_percent}
\end{equation}
with $0 <= \lambda_{min} <= \lambda_{max} <= 1$ being hyperparameters that control upper and lower percentage thresholds for the amount of $k$ to be kept. Furthermore, Eq.~\eqref{eq:part_bp_percent} considers $\lvert \varepsilon \rvert$ which is the difference between the current sample's loss and the maximum loss observed over the whole training so far. As a result, our approach allows for a dynamic update rate between $\lambda_{min}$ and $\lambda_{max}$ over the course of training, where the higher the training success becomes, i.e., the more the loss converges towards zero, the more the update rate will converge towards $\lambda_{min}$.

%Intuitively, this means that during early training, when the features have not yet been well trained by the DNN, a higher overall update rate per layer is chosen, while later during training, as the overall error size decreases, a lower update rate can be chosen. 

\begin{table}[t]
    \centering
    \caption{Datasets considered for transfer learning}
    \label{tab:transfer-datasets}
    \begin{tabular}{l c c c}
         \toprule
         \textbf{Dataset} & \textbf{Classes} & \textbf{Input Size} & \textbf{Type} \\
         \midrule
         cwru~\cite{cwrudataset} & 9 & $1\times512$ & Time Series \\
         daliac~\cite{leutheuser2013hierarchical} & 13 & $1\times1024$ & Time Series \\
         speech~\cite{warden2018speech} & 36 & $1\times16000$ & Time Series \\
         animals~\cite{animalsdataset} & 10 & $3\times128\times128$ & Vision \\
         cifar10~\cite{krizhevsky2009learning} & 10 & $3\times32\times32$ & Vision \\
         cifar100~\cite{krizhevsky2009learning} & 100 & $3\times32\times32$ & Vision \\
         flowers~\cite{nilsback2008automated} & 102 & $3\times128\times128$ & Vision \\
         \bottomrule
    \end{tabular}
\end{table}
\section{Evaluation}
\label{sec:evaluation}

We evaluate our approach on two training tasks: Transfer learning, see Secs.~\ref{sec:evaluation:fine-tuning},~\ref{sec:evaluation:across_mcus}, and~\ref{sec:evaluation:sparse-gradient}, and full DNN training, see Sec.~\ref{sec:evaluation:e2e}. We consider three DNN configurations for all experiments: \textit{uint8} where the complete DNN is fully quantized, \textit{mixed} where the DNN is quantized except its classification head that remains in floating-point, and \textit{float32} where the DNN remains completely in its original floating-point representation as a reference. Since we observed in our experiments that especially the quantization of tensors in the last layers of the DNN can significantly affect on-device training, we chose the mixed configurations as another trade-off between accuracy and resource utilization besides FQT. For all three DNN configurations, we used the micromod\footnote{\url{https://www.sparkfun.com/micromod}} platform and considered three MCUs for testing: IMXRT2062, nrf52840, and RP2040, see Tab.~\ref{tab:mcus} for detailed information. 
%We use the micromod platform and the Nordic Power Profiler Kit II to perform and measure all our experiments.

\subsection{On-Device Transfer Learning}
\label{sec:evaluation:fine-tuning}

\begin{figure*}[t]
    \centering
    \subfloat[Accuracy]{\label{fig:retrain_results:accuracy}%
        \includegraphics[width=0.48\textwidth]{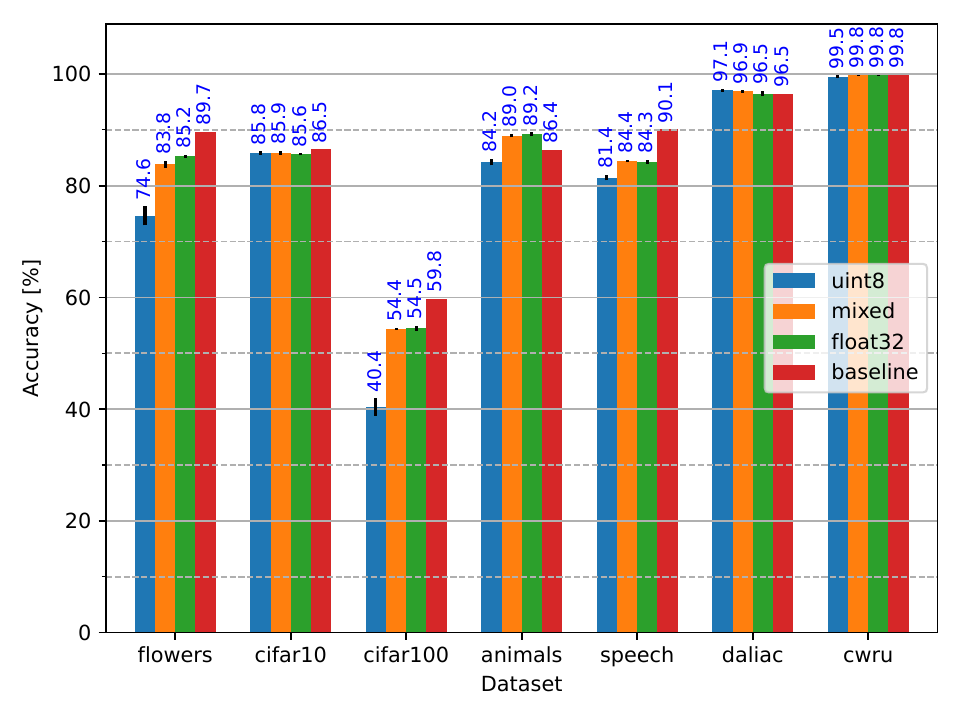}}
    \hfill
    \subfloat[Latency on IMXRT2062]{\label{fig:retrain_results:latency}%
        \includegraphics[width=0.48\textwidth]{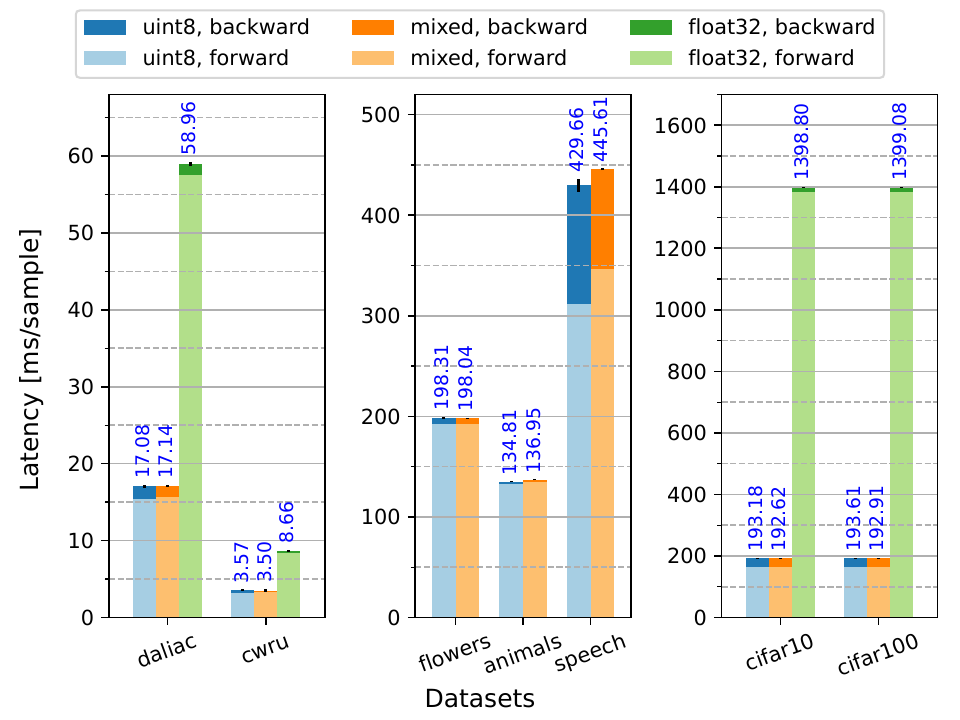}}
    \\
    \subfloat[Feature Maps, Trainable Weights, and Gradient Buffers]{\label{fig:retrain_results:memory_dynamic}%
        \includegraphics[width=0.48\textwidth]{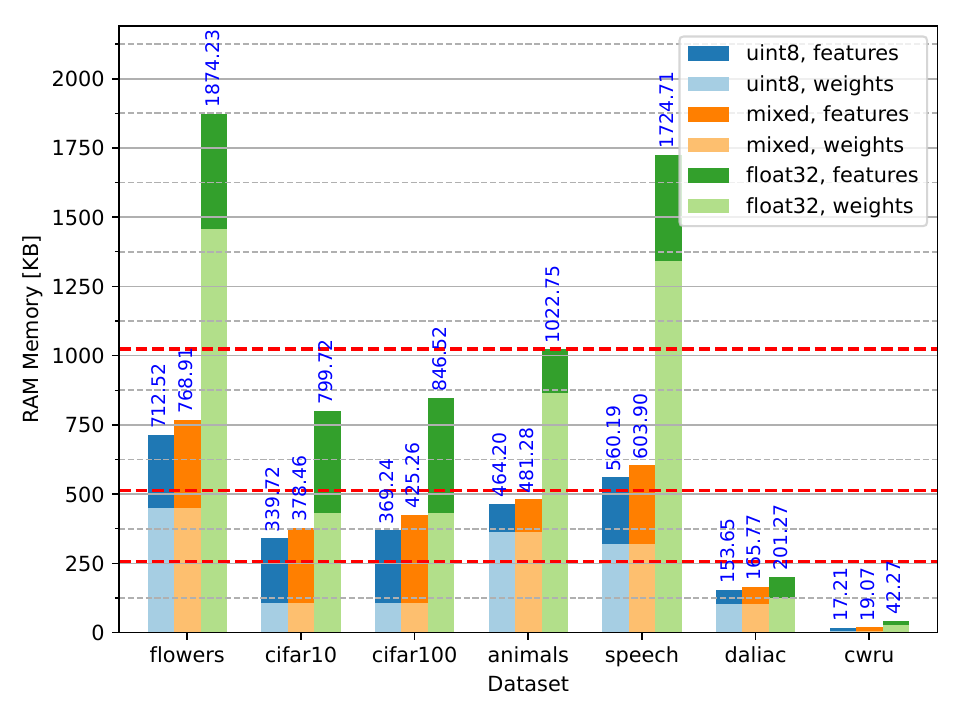}}
    \hfill
    \subfloat[Static Weights]{\label{fig:retrain_results:memory_static}%
        \includegraphics[width=0.48\textwidth]{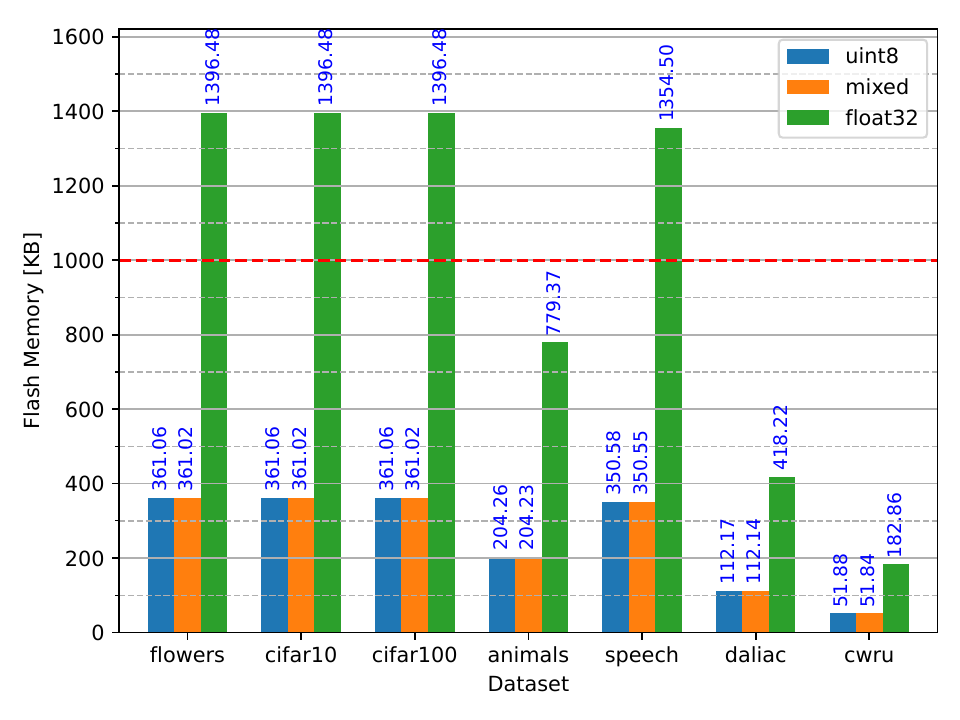}}
    \caption{Results of fully quantized on-device transfer learning (blue) compared to floating-point (green) and mixed training (orange). The accuracy results in Fig.~\ref{fig:retrain_results:accuracy} are averaged over five training runs. We also show baseline results trained on a GPU-based server in red. The latency results in Fig.~\ref{fig:retrain_results:latency} were measured on the IMXRT2062 MCU and are averaged over 1000 consecutive training steps. In both plots, we denote the standard deviation with black bars. Fig.~\ref{fig:retrain_results:memory_dynamic} and~\ref{fig:retrain_results:memory_static} show the memory utilization for Flash and RAM for all datasets as returned by our deployment framework. We have marked the memory constraints of the different MCUs with red dashed lines where relevant, see also Tab.~\ref{tab:mcus}.}
    \label{fig:retrain_results}
\end{figure*}

To validate the effectiveness of our on-device FQT approach, we performed transfer learning on seven different datasets, see Tab.~\ref{tab:transfer-datasets} for a detailed overview. 
%We report results for three time series datasets: cwru, daliac, and speech, and for four image classification tasks: animals, cifar10, cifar100, and flowers. 
We chose problems of different complexity, input size and domains to show the versatility of our approach in enabling stable FQT.

We use the same DNN architecture for all seven datasets which we dubbed \textit{MbedNet}. Its architecture is closely related to MobileNetV3~\cite{howard2019searching} scaled down to be more suitable to the hardware constraints on MCUs. To facilitate time series data, we map the time dimension of input samples to one of the spatial input dimensions of the DNN while leaving the other spatial dimensions empty. This way, MbedNet can support non-image input with only little changes to its architecture.

\begin{table*}
    \centering
    \caption{MCUs considered for our experiments}
    \label{tab:mcus}
    \begin{tabular}{l l c c c c}
    \toprule
    \textbf{Name} & \textbf{Type} & \textbf{Clock Speed} & \textbf{Amps Draw Idle} & \textbf{Flash/ROM} & \textbf{RAM}  \\
    \midrule
    RP2040 & Cortex-M0+ & 133 MHz & 31.24 mA & 16 MB (external) & 264 KB \\
    nrf52840 & Cortex-M4 & 64 MHz & 7.27 mA & 1 MB (internal) & 256 KB \\
    IMXRT2062 & Cortex-M7 & 600 MHz & 108.26 mA & 16 MB (external) & 2$\times$512 KB \\
    \bottomrule
    \end{tabular}
\end{table*}

We conducted our experiments for three DNN configurations (uint8, mixed, and float). First, we trained a baseline model on ImageNet for all vision datasets. We then performed transfer learning on each of the target datasets, including network pruning to adapt the feature extraction of the model. \blue{We used an iterative pruning schedule based on an $L1$-norm heuristic and automated gradual pruning.} Once trained, we performed post-training quantization for both FQT and mixed training before deploying the models on the MCU. Once deployed, we set the last five layers of each DNN to random values, thereby resetting its classification capabilities. \blue{We then retrained the DNNs using the original training datasets for 20 epochs with a learning rate of $0.001$ and a batch size of $48$.}

We report the results for all three DNN configurations and all tested datasets in Fig.~\ref{fig:retrain_results}. We show the accuracy on the test split after 20 epochs of retraining on-device in Fig.~\ref{fig:retrain_results:accuracy} (blue, orange, and green bars), averaged over 5 independent training runs and additionally provide the initial accuracy achieved by the model after GPU training (red bar, baseline).
In cases where on-device training was not possible due to memory constraints, we simulated retraining on a desktop computer\footnote{We used the same code and framework as used on the MCU minus some instruction set specific optimizations, which we had to disable.}. For on-device training, we did not implement any data augmentation on the MCU, which for some datasets, e.g., speech or flowers, resulted in more pronounced overfitting effects on device compared to the original GPU-based training, which included data augmentation.

For "simpler" problems with fewer classes, i.e., cwru, daliac, animals, and cifar10, we observed that on-device training was able to retrain or even improve the accuracy established by the baseline consistently and with almost no variance, while for "complex" problems with more classes, i.e., flowers, cifar100, speech, retraining slightly underperformed compared to the other DNN configurations and the baseline (compare blue bars with the other bars). Due to the reduced quantized range of the 8-bit integer gradient tensors, ambiguous or subtle features may not be correctly learned or discriminated, resulting in less precise decision boundaries between classes. For all datasets, this effect was easily resolved by training the classification head in floating-point, which resulted in a similar accuracy compared to full floating-point training (compare green and orange bars).

In Figs.~\ref{fig:retrain_results:memory_dynamic}, and~\ref{fig:retrain_results:memory_static}, we give an overview of the on-device memory utilization for each of the seven datasets for transfer learning. To execute DNNs on-device, our training framework requires three separate segments of memory. Two segments in RAM, see Fig.~\ref{fig:retrain_results:memory_dynamic} and one in Flash, see Fig.~\ref{fig:retrain_results:memory_static}. The segments in RAM are required to (a) store feature maps, i.e., intermediate results of the DNN layers during both the forward and backward pass, and (b) to store trainable weights and gradient buffers, while the segment in Flash memory is required for non-trainable weights. Since memory is one of the biggest limiting factors on all our target MCUs, see Tab.~\ref{tab:mcus}, we have highlighted relevant memory constraints with red dashed lines both in Fig.~\ref{fig:retrain_results:memory_dynamic} and Fig.~\ref{fig:retrain_results:memory_static}.

The differences that can be observed in memory utilization of the DNNs between the datasets are (a) due to the different sample sizes of the datasets, resulting in differently sized input and feature tensors, see Tab.~\ref{tab:transfer-datasets} and (b) the pruning we applied to the DNNs for some of the datasets during their initial training, resulting in smaller weight tensors. Compared to performing DNN inference on the device, additional RAM is required for weights and gradient buffers. Furthermore, the memory required for feature tensors may also increase, since intermediate results of layers must be kept longer in memory to facilitate the backward pass, resulting in fewer opportunities to reuse memory for subsequent intermediate tensors. In addition to trainable weights that must be stored in RAM, any weight tensor that is not to be retrained on-device can be stored read-only. Flash usage is shown in Fig.~\ref{fig:retrain_results:memory_static}. We found that especially floating-point training could exceed the Flash of some MCUs, e.g., the nrf52840 in our experiments. In general, however, this was much less of a concern than managing RAM utilization.

In addition to memory, Fig.~\ref{fig:retrain_results:latency} shows the latency per training sample, i.e., the time between the DNN receiving an input, producing a prediction, and completing backpropagation, as a unit of time for all datasets and DNN configurations on the IMXRT2062 that we could deploy on the MCU given its hardware constraints. All results are averaged over 1000 consecutive training samples and broken down by time spent in the forward and backward pass (compare light and dark colors).
%We denote the fraction per training sample spent in the forward pass in light colors, while we denote the fraction spent in backpropagation and stochastic gradient descent in dark colors. 
%All results shown are averaged over 1000 consecutive training steps. 
We observe that most of the training time was spent on the forward pass for transfer learning. This seems counterintuitive at first, since backpropagation theoretically performs roughly twice as many computations as the forward pass. However, since we are performing transfer learning rather than full DNN training in our experiments, backpropagation does not need to be executed for all layers of the DNN, only the last few. In addition, MBedNet is designed to learn compact representations quickly, resulting in large, computationally expensive initial layers, but compact, cheap final layers, allowing for a deep DNN architecture despite given resource constraints on MCUs. As a result, the time spent computing forward through the expensive initial layers still dominates the computation of the cheap final layers, even with the added overhead of backpropagation and training.

In conclusion, FQT significantly reduces memory requirements, making on-device training on MCUs feasible. Fine-tuning the final layers comes at little additional latency cost, and even when quantized, full baseline accuracy can be retrained for most of the datasets we tested. Where fully quantized training performed subpar, using a mixed approach where the classification head is trained unquantized while convolutional layers are trained quantized consistently addressed this problem while still offering an excellent tradeoff between memory utilization and latency. 

\begin{figure*}[t]
    \centering
    \subfloat[Latency]{\label{fig:compare:latency}%
        \includegraphics[width=0.48\textwidth]{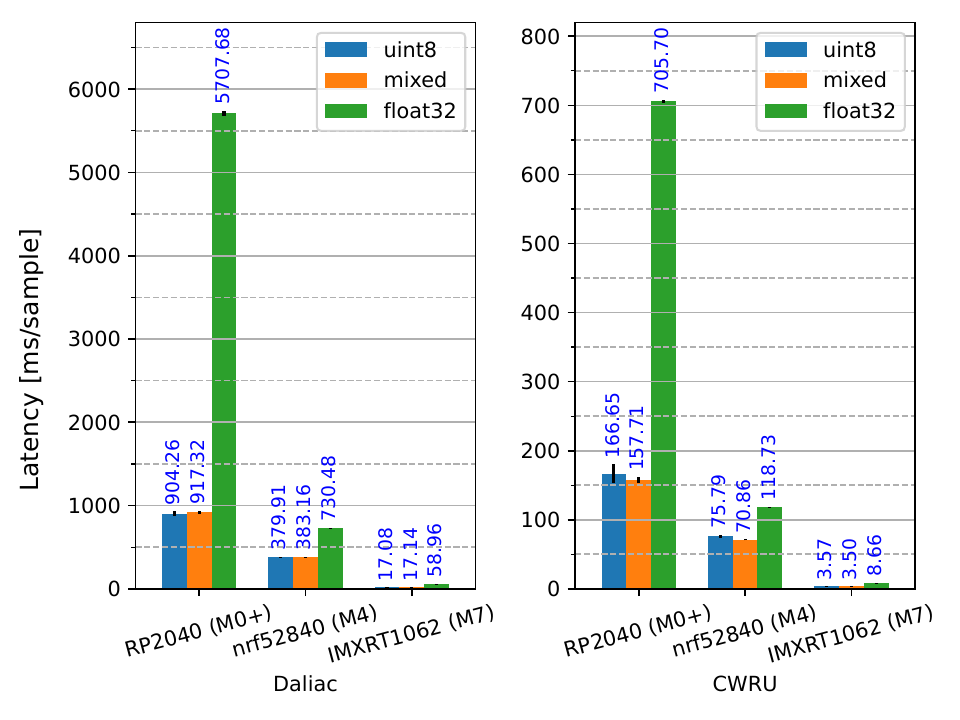}}
    \hfill
    \subfloat[Energy]{\label{fig:compare:energy}%
        \includegraphics[width=0.48\textwidth]{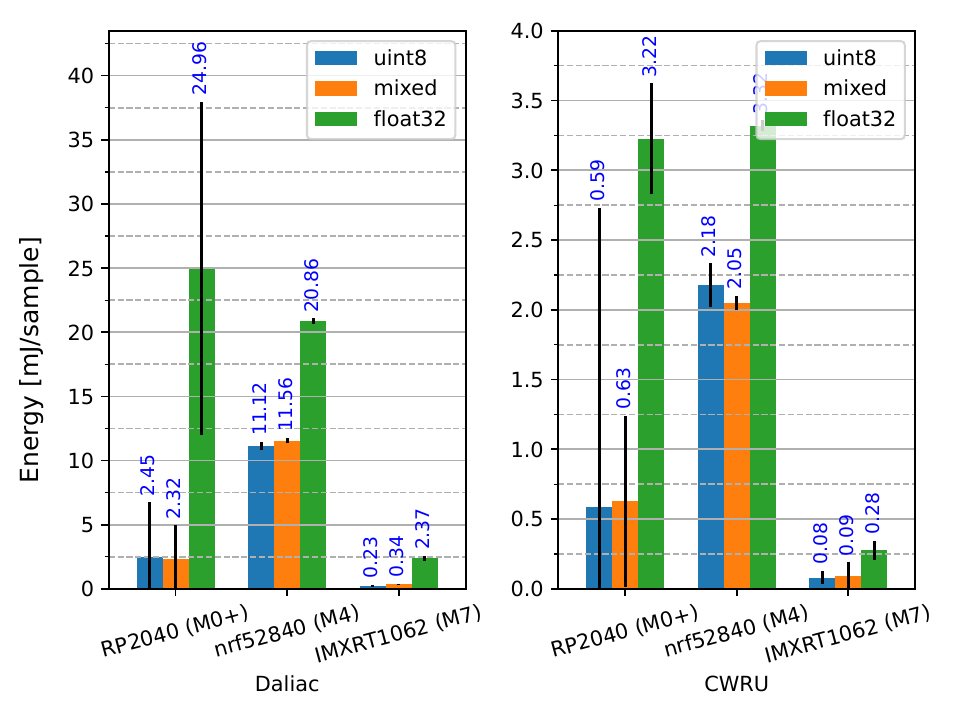}}
    \caption{Comparison of latency and energy for transfer learning on the CWRU and Daliac datasets for the three MCU platforms considered. All results are averaged over 1000 consecutive training steps, with the standard deviation shown as black bars.}
    \label{fig:compare}
\end{figure*}

\subsection{Transfer Learning Across MCUs}
\label{sec:evaluation:across_mcus}

As an additional experiment to the transfer learning task, and to get an understanding of the performance and constraints of FQT across different MCU platforms, we deployed the daliac and cwru datasets on the nrf52840 and RP2040 MCU and compared our findings with the results we observed on the IMXRT2062, see Fig.~\ref{fig:compare}. We only deployed the two datasets and not all seven because they were the only two that met the memory constraints on all three MCUs. As expected, processing on the IMXRT2062 is significantly faster than on the other Cortex-M4 and Cortex-M0+ based systems due to the much higher clock speed of the IMXRT2062, see Fig.~\ref{fig:compare:latency}. However, when comparing the RP2040 to the nrf52840 alone, even though the RP2040 has a higher clock speed, the nrf52840 was able to process training samples faster. This is because the nrf52840 has a dedicated floating-point unit (FPU) and implements the DSP instruction set extension, which our framework makes extensive use of to implement data-level parallelism (SIMD) in training sample processing, allowing much higher throughput than the RP2040, which has no FPU and no support for data-level parallelism.

In Fig.~\ref{fig:compare:energy}, we show the energy required per training sample averaged over 1000 consecutive training steps for each training configuration. Since we measured the entire MCU and not just the processor, we excluded the MCU's idle draw listed in Tab.~\ref{tab:mcus}. On a per-sample basis, the IMXRT2062 is the most energy-efficient and the NRF52840 is the least. However, this analysis assumes that samples can be processed as quickly as possible without delay between samples. In many real-world applications, new samples arrive at a fixed rate, slower than the MCU can process them, which can cause it to sit idle the rest of the time. As a result, in such scenarios, the MCU with the lowest idle power is usually the most energy-efficient, which according to Tab.~\ref{tab:mcus} would be the nrf52840 in our experiments.

\subsection{Dynamic Sparse Gradient Updates}
\label{sec:evaluation:sparse-gradient}

\begin{figure*}[t]
    \centering
    \subfloat[Accuracy float32]{\label{fig:partial:accuracy_float}%
        \includegraphics[width=0.48\textwidth]{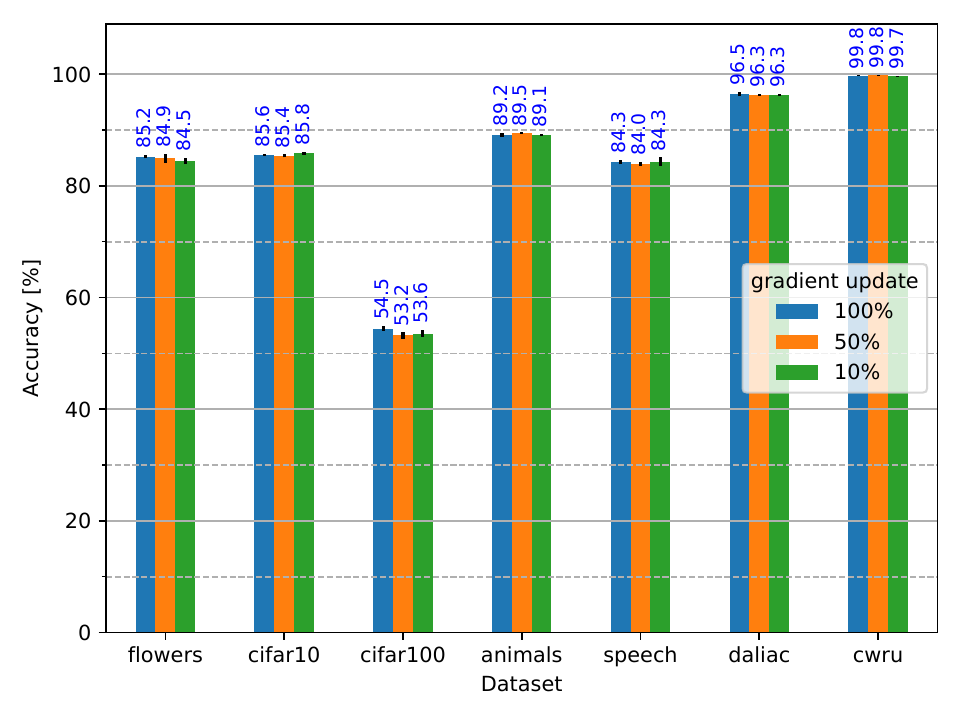}}
    \hfill
    \subfloat[Accuracy mixed]{\label{fig:partial:accuracy_mixed}%
        \includegraphics[width=0.48\textwidth]{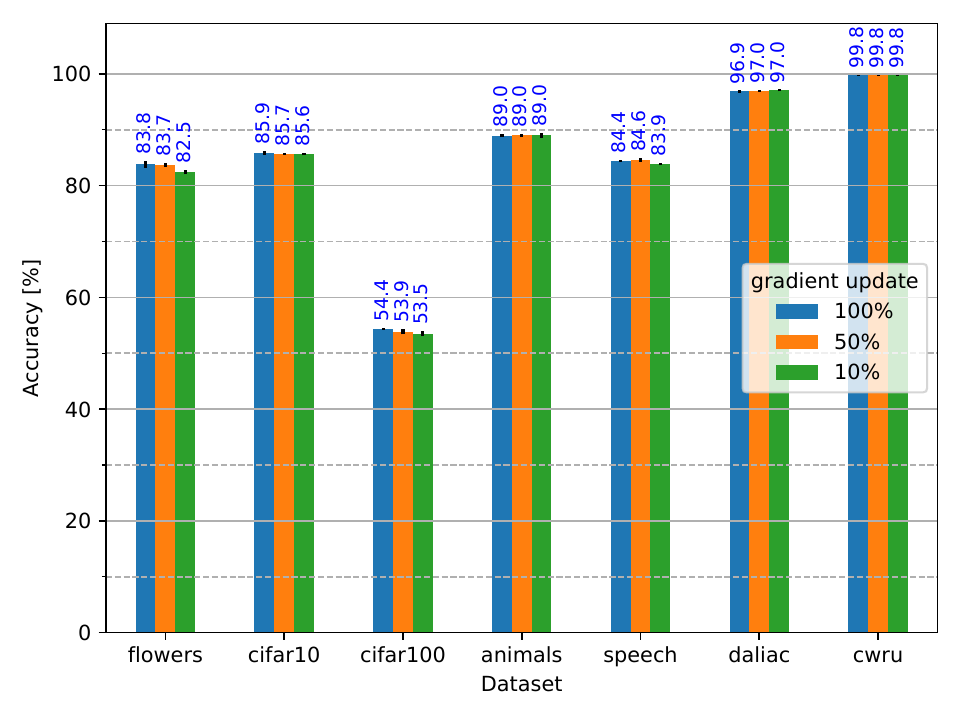}}
    \\
    \subfloat[Accuracy uint8]{\label{fig:partial:accuracy_uint8}%
        \includegraphics[width=0.48\textwidth]{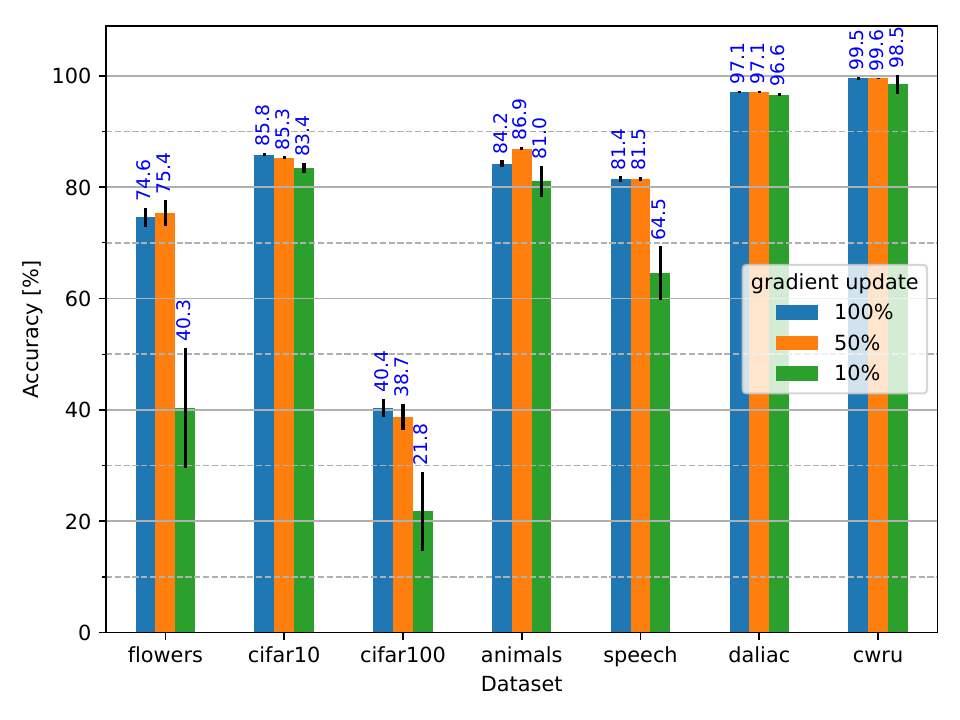}}
    \hfill
    \subfloat[Speedup per sample mixed on IMXRT2062]{\label{fig:partial:latency_mixed}%
        \includegraphics[width=0.48\textwidth]{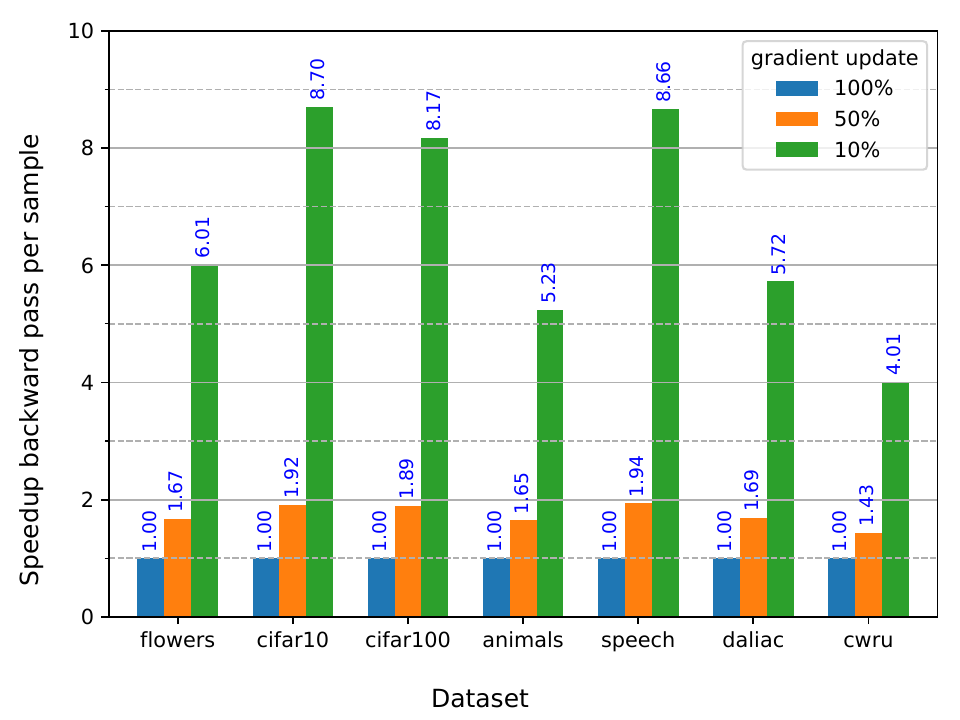}}
    \caption{Comparison of accuracy on the test datasets after 20 training epochs for three different gradient update rates ($\lambda_{min} \in \{0.1, 0.5, 1.0\}$) for the mixed, quantized, and floating-point DNN configurations in Figs.~\ref{fig:partial:accuracy_float} to~\ref{fig:partial:accuracy_uint8}. The results we report are averaged over 5 training runs. We show the standard deviation as black bars. Furthermore, we show the average speedup per sample for the different gradient update rates for the mixed training configuration in Fig.~\ref{fig:partial:latency_mixed}.}
    \label{fig:partial}
\end{figure*}

We validate the ability of our partial gradient updating approach to reduce the computational complexity of backpropagation by comparing the training accuracy achieved with different dynamic gradient update rate configurations, see Fig.~\ref{fig:partial}, with the results of our experiments presented in Section~\ref{sec:evaluation:fine-tuning}, where we used full gradient updates. We show the test accuracy for three different $\lambda_{min}$ (blue: $\lambda_{min}=1.0$, orange: $\lambda_{min}=0.5$, and green: $\lambda_{min}=0.1$) for all DNN configurations and for all seven datasets considered in our transfer learning experiments, see Figs.~\ref{fig:partial:accuracy_float},~\ref{fig:partial:accuracy_mixed}, and~\ref{fig:partial:accuracy_uint8}. For all experiments, we performed on-device training in combination with dynamic gradient updates for 20 epochs, and we repeated each training 5 times and provide averaged results.

For both floating-point and mixed transfer learning (Figs.~\ref{fig:partial:accuracy_float} and~\ref{fig:partial:accuracy_mixed}), both $\lambda_{min}=0.5$ and $\lambda_{min}=0.1$ achieved the same level of accuracy after 20 epochs as $\lambda_{min}=1.0$, showing that, if chosen appropriately, only a small fraction of gradients need to be actually updated for each training sample to achieve a similar training result as with fully updated gradients. For FQT (Fig.~\ref{fig:partial:accuracy_uint8}), $\lambda_{min}=0.5$ also worked without significant degradation compared to $\lambda_{min}=1.0$. However, with the extremely small update rate of $\lambda_{min}=0.1$, we observed that the training became unstable, i.e., high variance between different training runs, and performed significantly worse than $\lambda_{min}=1.0$. 

\begin{figure}
    \centering
    \includegraphics[width=.85\linewidth]{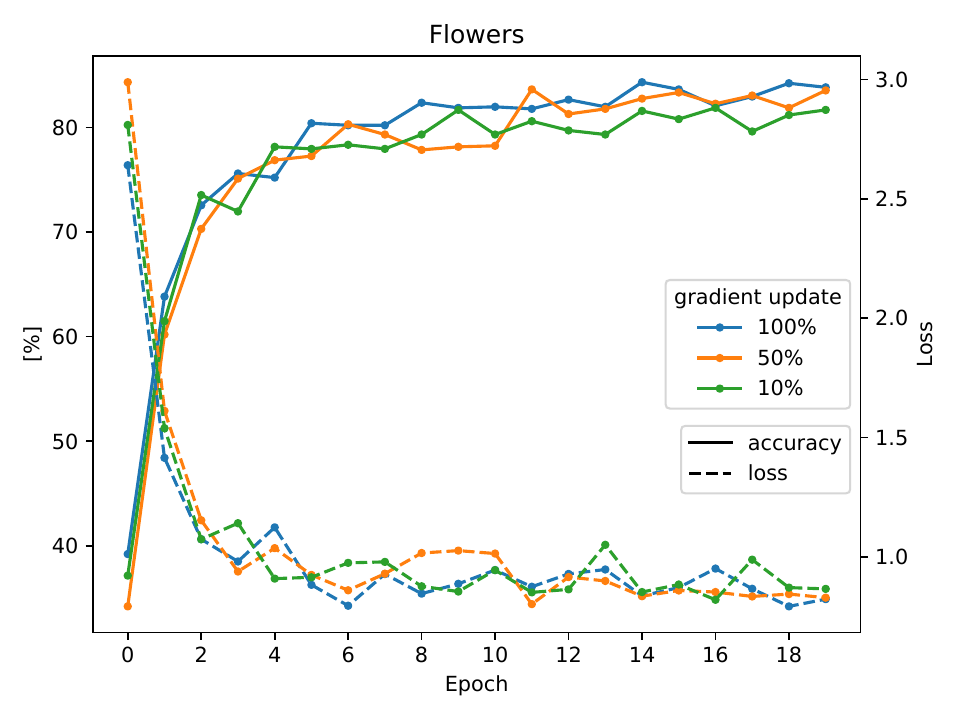}
    \caption{Accuracy and loss curves for all three considered gradient update rates, exemplarily for a training run on the flowers dataset using the mixed DNN configuration}
    \label{fig:partial:train_flowers_mixed}
\end{figure}

To further demonstrate that introducing partial gradient updates actually reduces the computational complexity of backpropagation and does not just result in a slower loss convergence over the course of training, we provide both loss and accuracy curves exemplarily for the flowers dataset and for the mixed DNN configuration for all three $\lambda_{min}$ tested in Fig.~\ref{fig:partial:train_flowers_mixed}. It can be seen that the loss convergence speed does not decrease for lower $\lambda_{min}$ and is relatively consistent across all three different gradient update rates, showing that dynamic gradient updating is able to reduce computational complexity while minimally impacting training performance.

Finally, we show the speedup achieved by partial gradient updating with all three $\lambda_{min}$ considered when using the mixed training configuration on the IMXRT2062 for all seven datasets, see Fig.~\ref{fig:partial:latency_mixed}. We observed that for the mixed DNN configuration, reducing the number of updated gradients leads to a significant speedup per sample for backpropagation, especially for $\lambda_{min}=0.1$, where an average speedup of about $6.64$ was achieved across all datasets. Since the overhead of partial gradient updating is the same for different $\lambda_{min}$, while at the same time, a smaller $\lambda_{min}$ leads to more operations being skipped, the increase in speedup is polynomial, depending on the number of loop nests in the layers trained.

\subsection{Complete On-Device Training}
\label{sec:evaluation:e2e}

\begin{figure*}[t]
    \centering
    \subfloat[Accuracy]{\label{fig:e2e:accuracy}%
        \includegraphics[width=0.48\textwidth]{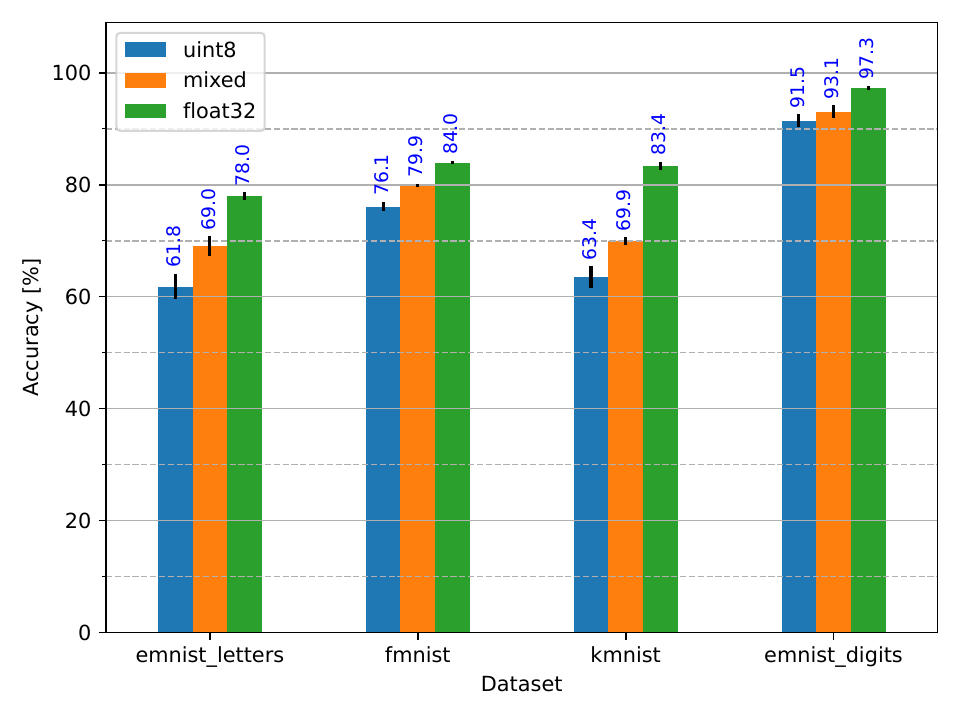}}
    \hfill
    \subfloat[Latency and Energy]{\label{fig:e2e:latency}%
        \includegraphics[width=0.48\textwidth]{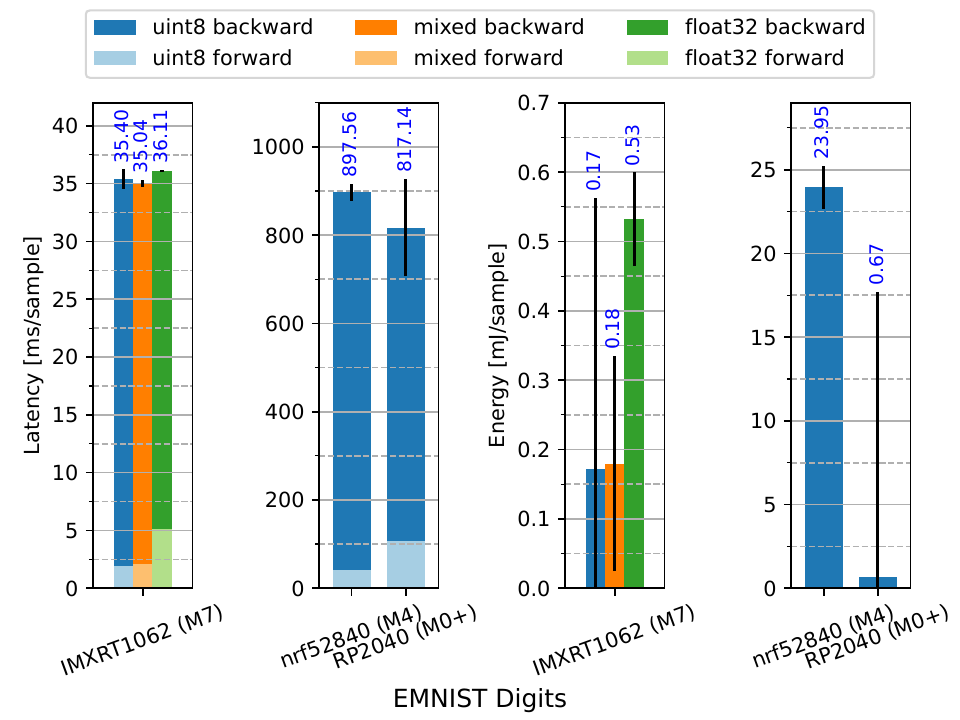}}
    \caption{Complete On-Device training of FMNIST, KMNIST, EMNIST Letters, and EMNIST Digits for a DNN consisting of two convolutional and two linear layers, both with ReLU and batch normalization, and with max-pooling in between. The accuracy results shown in Fig.~\ref{fig:e2e:accuracy} are averaged over 5 training runs, while the latency and energy results shown in Fig.~\ref{fig:e2e:latency} are exemplary for EMNIST for all three considered MCU platforms and are averaged over 1000 consecutive training steps. We denote the standard deviation with black bars.}
    \label{fig:e2e}
\end{figure*}

To evaluate that our FQT approach can also be used to train DNNs on MCUs completely from scratch, we used a smaller DNN pre-trained on the MNIST dataset~\cite{lecun1998gradient}, which we then fully retrained, i.e., all layers, on the device for four MNIST-related datasets, see Tab.~\ref{tab:datasets_e2e}. The DNN we used for this purpose consisted of 2 convolutional layers, a max-pooling layer, and 2 linear layers, all with ReLU as activation and BatchNorm.
%and pre-trained it on the MNIST dataset~\cite{lecun1998gradient}. 
%Then we performed on-device transfer learning of the complete DNN on four MNIST-related datasets, FMNIST~\cite{xiao2017fashion}, KMNIST~\cite{clanuwat2018deep}, and two versions of EMNIST~\cite{cohen2017emnist}, one version conatining the 10 digits 0 to 9 and the other version the 26 letters of the alphabet. 
\blue{We used the same DNN configurations as in Sec.~\ref{sec:evaluation:fine-tuning}, a learning rate of $0.001$ and a batch size of $48$. The corresponding results are shown in Fig.~\ref{fig:e2e}.}

\begin{table}
    \centering
    \caption{Datasets considered for full on-device training}
    \label{tab:datasets_e2e}
    \begin{tabular}{l c c c}
      \toprule
      \textbf{Dataset} & \textbf{Classes} & \textbf{Input Size} & \textbf{Type}  \\
      \midrule
      FMNIST~\cite{xiao2017fashion} & 10 & $1\times28\times28$ & Vision \\
      KMNIST~\cite{clanuwat2018deep} & 10 & $1\times28\times28$ & Vision \\
      EMNIST-Letters~\cite{cohen2017emnist} & 26 & $1\times28\times28$ & Vision \\
      EMNIST-Digits~\cite{cohen2017emnist} & 10 & $1\times28\times28$ & Vision \\
      \bottomrule
    \end{tabular}
\end{table}

In Fig.~\ref{fig:e2e:accuracy}, we show the accuracy results after on-device training for all three DNN configurations and for all four datasets. On-device full floating-point training achieved the highest accuracy, closely followed by mixed and fully quantized training. Different from transfer learning, we observed a more pronounced gap between quantized and floating-point training. A reason for this observation is, that while the transfer learning experiments we discussed earlier only retrained the classification head and the last convolutional layers of the DNN, and thus could still rely on a well-trained feature extractor, i.e., the earlier layers, the full on-device experiments shown in Fig.~\ref{fig:e2e:accuracy} had to learn both feature extraction and classification from scratch. As a result, it appears that the limited resolution of the 8-bit data types used for training resulted in the DNN not being able to adequately learn the extraction of more complex features in its earlier layers. Therefore, although the training was still generally successful for FQT, we observed an earlier stagnation of FQT compared to floating-point or mixed training, and thus a slightly reduced accuracy. Similar effects can be observed with training of other kinds of low-bit representation DNNs like binarized DNNs~\cite{hubara2016binarized}. A more fine-grained quantization scheme, such as per-channel quantization~\cite{krishnamoorthi2018quantizing}, or alternative quantization strategies, such as weighted entropy~\cite{park2017weighted} or stochastic quantization~\cite{Dong2019StochasticQF}, while not as well supported by existing tool flows on MCUs, could be useful approaches to bridge this gap as long as they can maintain a reasonable computational overhead. %We consider the exploration of such trade-offs as future work.

We also present latency and energy results for our complete on-device training experiments exemplarily for the EMNIST Digits dataset and for all three considered MCU platforms and training configurations in Fig.~\ref{fig:e2e:latency}. For the nrf52840 and RP2040, we were only able to test the uint8 configuration due to memory constraints on both platforms. 
As in Fig.~\ref{fig:retrain_results:latency} above, we mark the fraction of latency per sample spent in the forward pass in light colors and in the backward pass in dark colors. Since we are training all layers of the DNN, the time spent in the backward pass is significantly higher than the time spent in the forward pass. These observations are the opposite of what we saw earlier in Fig.~\ref{fig:retrain_results:latency}, where only the last layers were trained, but they are much closer to our initial intuition based on the theoretical analysis of FQT training that we stated at the beginning of the paper. Nevertheless, the results presented in Fig.~\ref{fig:e2e:latency} clearly show the significant computational impact that especially training of early feature extracting layers has.  
%However, the relationship seen in Fig.~\ref{fig:e2e:latency} is more in line with what we would expect to see based on a theoretical analysis of our FQT training algorithm. 

Finally, we show energy results in Fig.~\ref{fig:e2e:latency}. Identical to the results discussed in Fig.~\ref{fig:compare:energy}, we subtracted the idle draw of the MCUs shown in Tab.~\ref{tab:mcus}. As with the results discussed earlier, we can see that the IMXRT1062 is the most energy-efficient option per sample compared to the nrf52840 and RP2040, which instead have a much lower idle power draw.

\blue{\subsection{Comparison with MCUNet-5FPS}}
\label{sec:evaluation:mcunet}

\begin{table*}[t]
    \blue{
    \centering
    \caption{Comparison between our quantized optimizer and SGD+M+QAS~\cite{lin2022device} for updating the last two blocks of MCUNet-5FPS~\cite{lin2020mcunet}. For our approach, we provide the accuracy on the test splits of the datasets.}
    \label{tab:mcunet}
    \begin{tabular}{ l l c c c c c c c c c}
    \toprule
    \multirow{2}{*}{Precision} & \multirow{2}{*}{Optimizer} & \multicolumn{8}{c}{Accuracy (\%) (MCUNet backbone: 23M MACs, 0.48M Param)} & \multirow{2}{*}{\shortstack{Avg.\\Acc.}} \\
    \cmidrule(lr){3-10} 
    & & Cars~\cite{krause20133d} & CF10~\cite{krizhevsky2009learning} & CF100~\cite{krizhevsky2009learning} & CUB~\cite{welinder2010caltech} & Flowers~\cite{nilsback2008automated} & Food~\cite{bossard2014food} & Pets~\cite{parkhi2012cats} & VWW~\cite{chowdhery2019visual} & \\
    \midrule
    \texttt{fp32} & SGD-M~\cite{lin2022device} & 56.7 & 86.0 & 63.4 & 56.2 & 88.8 & 67.1 & 79.5 & 88.7 & 73.3 \\
    \midrule
    \texttt{int8} & SGD-M~\cite{lin2022device} & 31.2 & 75.4 & 64.5 & 55.1 & 84.5 & 52.5 & 79.5 & 88.7 & 64.9 \\ 
    \texttt{int8} & SGD+M+QAS~\cite{lin2022device} & \textbf{55.2} & 86.9 & 64.6 & 57.8 & \textbf{89.1} & 64.4 & \textbf{80.9}      & 89.3 & 73.5 \\ 
    \midrule
    \texttt{uint8} & \textbf{ours} & 54.5 & \textbf{89.5} & \textbf{65.2} & \textbf{58.5} & 85.8 & \textbf{66.6} & 79.8 & \textbf{89.3} & \textbf{73.7} \\
    \bottomrule
    \end{tabular}}
\end{table*}

\blue{We compare the performance of our FQT optimizer with the results reported by the authors of the optimizer SGD+M+QAS~\cite{lin2022device} when training the last two blocks of MCUNet-5FPS~\cite{lin2020mcunet}, see Tab.~\ref{tab:mcunet}. The first row in Tab.~\ref{tab:mcunet} shows the baseline accuracy reported by~\cite{lin2022device} for MCUNet trained on eight datasets using a floating-point SGD optimizer with momentum (SGD-M). We could reproduce these results using the publicly available sources of MCUNet provided by the authors. The next two rows show the results obtained by~\cite{lin2022device} first for naively using quantized SGD-M without any changes besides the usage of quantized tensors and second for their SGD+M+QAS approach, which addresses shortcomings of the SGD-M approach. The last row shows the results we obtained for retraining MCUNet using our approach, i.e., fully quantized training with standardized gradients and dynamic adaptation of the zero-point and scale parameters as described in Sec.~\ref{sec:implementation}. All our reported results are on unseen test splits of the datasets. Like~\cite{lin2022device} we retrained for 50 epochs. We used a learning rate of $0.001$ and a batch size of $48$.}

\blue{Our optimizer matched SGD+M+QAS closely, even outperforming it in five of the eight datasets, see the last two rows in Table~\ref{tab:mcunet}. On average, this resulted in our approach having a slightly higher performance in retraining MCUNet, see last column of Table~\ref{tab:mcunet}, while exceeding the performance of naively retraining the quantized weights, i.e. SGD-M.}

\begin{figure}
    \blue{
    \centering
    \includegraphics[width=.85\linewidth]{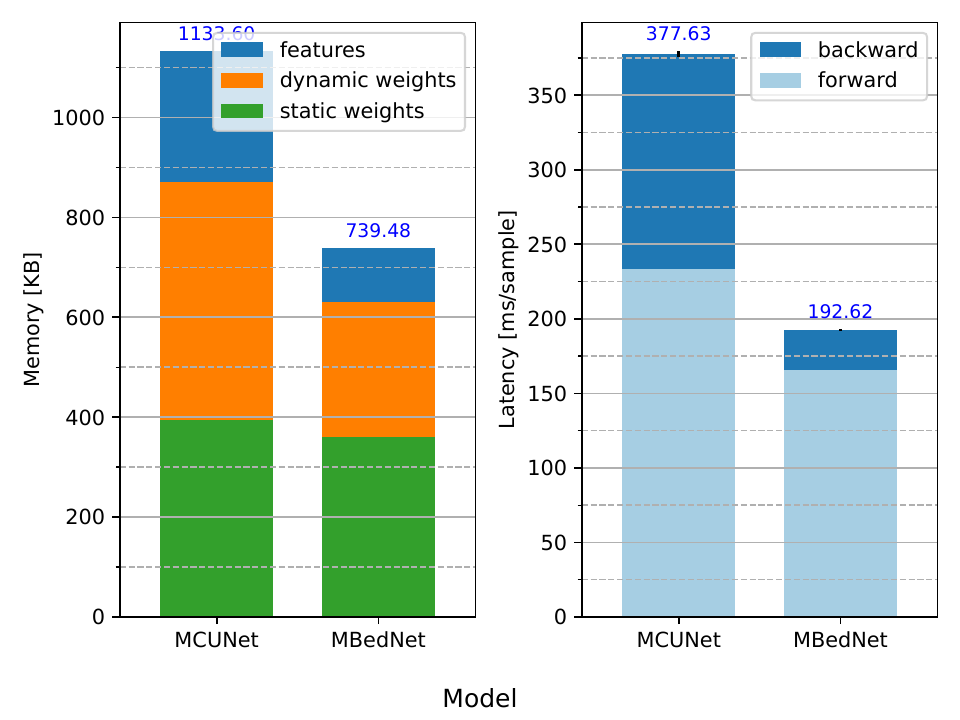}
    \caption{Memory (green: ROM, orange + blue: RAM) and latency averaged over 1000 consecutive training steps of MBedNet (ours) compared to MCUNet-5FPS deployed on the IMXRT2062 for cifar10.}
    \label{fig:mcunet_vs_mbednet}}
\end{figure}

\blue{However, the main goal of both our optimizer and SGD+M+QAS is to provide a better tradeoff between memory and latency to meet the resource constraints of MCUs and enable on-device training. But since the authors of~\cite{lin2022device} do not report detailed numbers for SGD+M+QAS, we instead report a comparison between MCUNet deployed with our framework and MbedNet for cifar10, see Fig.~\ref{fig:mcunet_vs_mbednet}.}

\blue{MCUNet has more trainable parameters than MBedNet, especially in its last layers. This results in (a) 34.8\% less memory requirements and (b) 49.0\% smaller latency of MBedNet compared to MCUNet, see Fig.~\ref{fig:mcunet_vs_mbednet}. For memory, this results in (a) larger feature maps to propagate between layers (blue segment of bars), and (b) more memory required to store larger gradient and weight buffers to enable training of the last layers (orange segment of bars). This trend can likewise be observed for latency, see the left plot in Fig.~\ref{fig:mcunet_vs_mbednet}. Because MBedNet's final layers are smaller, both their forward and backward passes can be executed significantly faster than MCUNet's. Since the backward pass has to compute two partial derivatives for each layer, the difference is more pronounced than for the forward pass. Finally, when comparing accuracy, both architectures achieve a similar baseline of around 86\%. When  using our optimizer, MCUNet could be trained to an accuracy of 89.5\% for cifar10, which is slightly higher than the 85.9\% our optimizer achieved for MBedNet but which is in line with what SGD-M-QAS achieved on MCUNet (86.9\%). Still, we argue that the significantly lower memory and latency cost of MBedNet compared to MCUNet makes it a more suitable architecture for efficient on-device training on MCUs. This allows MbedNet not only to run inference at around 5FPS (200 ms) like MCUNet, but also to retrain.}

\section{Conclusion}
\label{sec:conclusion}

In this work, we discussed the challenges of on-device training on Cortex-M microcontrollers, such as computational overhead and memory constraints, and proposed a method to overcome them. To this end, we proposed a combination of fully quantized on-device training (FQT) combined with dynamic partial gradient updates to reduce compute overhead by dynamically adapting to the distribution and magnitude of gradients values during backpropagation. In an extensive evaluation, focusing on three different Cortex-M-based MCUs, we show that on-device transfer learning and full on-device training are both feasible from an accuracy and resource utilization standpoint. Therefore, our results indicate that on-device MCU training is possible for a wide range of datasets.
Furthermore, our method is flexible and allows for training configurations, where some layers of a DNN are trained in floating-point, while others are trained quantized. This solidifies the capability of our method in adapting to problems of varying complexity and to different DNN architectures.

\bibliographystyle{IEEEtran}
\bibliography{references}

\end{document}